\documentclass[
]{ceurart}

\usepackage{listings}
\usepackage{graphicx}
\usepackage{subcaption}
\usepackage{url}
\makeatletter
\g@addto@macro{\UrlBreaks}{\do\/\do\-}
\makeatother
\begin{document}

\copyrightyear{2025}
\copyrightclause{CEUR requirements: “Copyright © 2025 for this paper by its authors. Use permitted under Creative Commons License Attribution 4.0 International (CC BY 4.0)5}

\conference{Proceedings of the Joint Ontology Workshops (JOWO) - Episode XI: The Sicilian Summer under the Etna, co-located with the 15th International Conference on Formal Ontology in Information Systems (FOIS 2025), September 8-9, 2025, Catania, Italy}

\title{Meanings are like Onions: a Layered Approach to Metaphor Processing}

\tnotemark[1]

\author[1]{Silvia Cappa}
\ead{silviacappa@cnr.it}
\address[1]{Institute for Cognitive Sciences and Technologies (ISTC), CNR, Rome, Italy}

\author[1,2]{Anna Sofia Lippolis}
\ead{annasofia.lippolis2@unibo.it}
\address[2]{University of Bologna, Italy}

\author[3]{Stefano Zoia}
\ead{stefano.zoia@unito.it}
\address[3]{University of Turin, Italy}

\fntext[1]{Equal contribution.}


\begin{abstract}
  Metaphorical meaning is not a flat mapping between concepts, but a complex cognitive phenomenon that integrates multiple levels of interpretation. In this paper, we propose a stratified model of metaphor processing that treats meaning as an onion: a multi-layered structure comprising (1) contextual information, (2) conceptual blending analysis, and (3) pragmatic analysis. This three-dimensional framework allows for a richer and more cognitively grounded approach to metaphor interpretation in computational systems. At the first level, metaphors are annotated through contextual metadata. At the second level, we model conceptual combinations, linking components to emergent meanings. Finally, at the third level, we introduce a pragmatic vocabulary to capture speaker intent, communicative function, and contextual effects, aligning metaphor understanding with pragmatic theories. By unifying these layers into a single formal framework, our model lays the groundwork for computational methods capable of representing metaphorical meaning beyond surface associations—toward deeper, more context-sensitive reasoning.
\end{abstract}


\maketitle

\section{Introduction}
\label{sec:intro}
Metaphors pervade human communication and cognition, extending far beyond mere linguistic decoration. As cognitive tools, they grant privileged access to implicit knowledge structures that might otherwise remain hidden \cite{moser2004role}. By mapping relationships between concepts, metaphors serve as bridges that both reveal and reshape our conceptual frameworks—think of how we say that we \textit{spend}, \textit{save}, or \textit{waste} time, implicitly assuming it is a finite resource. 

Despite rapid progress in natural language processing, computational metaphor analysis continues to face five intertwined challenges rooted in the very knowledge structures metaphors invoke:
\begin{enumerate}
    \item \textit{Data scarcity and representational gaps}. Datasets accounting for many metaphorical phenomena are scarce, and building new ones is (i) resource-intensive, and (ii) hindered by frameworks that go no further than simple domain mappings.
    \item \textit{Contextual insensitivity}. While the Conceptual Metaphor Theory (CMT) developed by Lakoff and Johnson \cite{lakoff1981} dominates in computational accounts of metaphors, it often fails to capture, among other things, how context shifts a metaphor’s meaning in discourse.
    \item \textit{Evaluation and standardization}. Definitions, metrics, and supported linguistic forms (nominal, verbal, adjectival) in metaphor processing vary wildly across studies \cite{hicke2024science}.
    \item \textit{Theoretical fragmentation}. Competing accounts (e.g., CMT vs. interactional or embodiment theories) illuminate different aspects of metaphorical phenomena but rarely integrate pragmatics—what a metaphor does in conversation—often goes unmodeled despite Speech Act Theory (SAT) being a long-standing account of these processes and widely adopted in the literature \cite{dancygier2016figurativeness}.
    \item \textit{Limits of current computational works}. Even large language models (LLMs), the state of the art in metaphor processing, struggle to distinguish deep relational mappings from mere associations, particularly in complex or multimodal metaphors \cite{wijesiriwardene2023analogical, nezhurina2024alice}.
\end{enumerate}

Given this context, in this paper we aim to articulate a theoretical proposal that may serve both as a fruitful direction for future research and as a foundation for subsequent empirical work on this issue. Specifically, we put forward an operational framework designed to support the processing of metaphorical meaning in a way that can effectively represent and interweave both conceptual and pragmatic aspects.

The starting point for this framework lies in the observation that, for analytical purposes, implicit linguistic meaning can be treated as operating on a different level from that of explicitly encoded, surface-level meaning. Conceptual and pragmatic meanings are not directly encoded in the expression itself; rather, they are inferred through processes of contextualization and implicature.
Consider, for example, the sentence \textit{we are wasting our time}: the conceptual interpretation—where notions such as finite resources, waste, and time emerge, and where an analogical operation groups features of these concepts—is not directly expressed by the literal sentence. Similarly, the utterance’s potential pragmatic function—i.e., the way that the utterance is intended—is also not explicitly encoded.
This distinction reflects a view of meaning as a multi-level phenomenon—not necessarily intended as a model of how language inherently works, but as a potentially effective way of structuring meaning for computational processing, where explicitly encoded meaning and interpretation can be distinguished from conceptual and pragmatic ones, which remain implicit. Although stratified or hierarchically structured models of meaning are well established in the literature on Pragmatics (e.g. in Grice \cite{grice1971meaning}), computational metaphor analysis and pragmatics have yet to establish a systematic connection—a link that could substantially advance research in both domains.

Therefore, from this theoretical perspective we propose a framework focused on the implicit level of meaning that aims to study metaphors as \textit{cognitive tools in context}, which do not exist as an abstract operation, separate from its usage or from concrete linguistic experience. A simultaneous cognitive and pragmatic analysis would in fact support the view of metaphor as a tool that effectively works—that is, it achieves communicative efficacy—when it is perceived as conceptually appropriate and functionally aligned with the speaker’s pragmatic goal, in addition to serving as models that reflect the cognitive structuring of experience, shaping both reasoning and action \cite{moser2004role}. In this view, the implicit knowledge conveyed by metaphors can be processed more fruitfully in computational systems if conceived as a multi-layered entity, like an onion. Each of our onion layer can be ontologically represented, providing a structured and functional approach to modeling metaphorical knowledge that aligns with our analytical goals. If the outermost layer correspond to the level of contextual information mapping, moving inward reveals the level at which the analogical conceptual operations take place, and deeper still, the pragmatic intention that motivated the entire utterance. By connecting CMT’s cognitive account of metaphor with SAT–inspired pragmatics, our stratified framework aim to capture both what metaphors map and what they accomplish in communication. This unified model lays the groundwork for richer datasets, standardized evaluation, and computational systems that better mirror how humans unlock implicit knowledge through metaphor. 

The remainder of this paper is structured as follows. In Section~\ref{sec:related}, we describe the related works. Section~\ref{sec:proposal} introduces our three‐dimensional model of meaning processing, illustrating how conceptual and contextual layers can be systematically integrated. In Section~\ref{sec:implications}, we explore the consequences of this multi‐layered view for computational processing of metaphors, discussing both implementation strategies and evaluation methods. Finally, Section~\ref{sec:conclusion} summarizes our contributions and outlines directions for future work.

\section{Related Work}
\label{sec:related}
In this section, we describe the theorical background for our proposal and discuss the challenges related to existing works on metaphor processing.

\subsection{Metaphor theories}
CMT, developed by Lakoff and Johnson, posits that metaphors map a source domain onto a target domain via systematic correspondences, enabling abstract reasoning through familiar experiential structures~\cite{lakoff1981}. Conceptual Blending Theory (CBT) extends this view by introducing a generic ``blend'' space that selectively inherits elements from both input domains according to a blending criterion or key property, such as \emph{yellow} when we say ``golden hair'', evoking that golden is yellow and shiny like the hair~\cite{fauconnier2003conceptual}. CBT is regarded as a valid computational approach also by the Categorization theory \cite{holyoak2018metaphor}. This account sees metaphors as category statements where the source acquires a categorical meaning, more abstract than its literal meaning. For example, ``golden'' in ``golden hair'' denotes the category of ``shiny, yellow things''. In this view, conceptual blending can be used to extract the abstract meaning of the source and combine it with the meaning of the target. These frameworks emphasize that metaphor comprehension relies on shared background knowledge (or frames), which Fillmore’s frame semantics formalizes by associating lexical items with structured role–filler expectations~\cite{fillmore_06}. The complementary relationship between CMT, CBT, and frame semantics highlights that metaphorical meaning emerges not merely from lexical similarity but from dynamic frame activation and role alignment within a community’s commonsense knowledge~\cite{dancygier2016figurativeness}.
Beyond commonsense or prototypical knowledge, recent theories of metaphors have noted the lack of inclusion of personal and sometimes contextual aspects that influence knowledge acquisition and interchange.
In fact, the experiential dimension of metaphor has traditionally been downplayed, with research focusing primarily on metaphors as a mental and individual achievement. Researchers have so far paid little attention to context and the collaborative production of metaphoric language~\cite{jensen2018world}.

The communicative aspect of metaphor is fundamental to view metaphor as a multidimensional phenomenon \cite{semino2008metaphor}. Metaphor systems are not neutral but reflect underlying belief systems that justify social actions and representations. In this view, language and metaphor in particular plays a key role in realizing these social and political values: texts are always ``oriented social action'' \cite{titchkosky2007reading}.
Linell's notion of an ``interworld'' provides a valuable theoretical framework for understanding these social dimensions of metaphor \cite{linell2009rethinking}. Unlike traditional cognitive approaches that locate metaphor primarily in individual minds, the interworld concept emphasizes how metaphorical meaning emerges through interaction in a shared communicative space. 
As an example, long-standing views of metaphor like the one carried out by CMT presuppose universal bodily experiences, excluding experiences of the disabled~\cite{schalk2013metaphorically}. For this reason, recent studies claim for a view on metaphor that is not just embodied, but inter-bodily. Indeed, Gibbs~\cite{gibbs2017metaphor}, contrary to the standard assumption within CMT that claims source domains of conceptual metaphors are primarily based on direct sensorymotor experiences, argues that metaphorical meanings do not necessarily arise from the mappings of purely embodied knowledge onto abstract concepts. Instead, the source domains themselves metaphorical in nature. 

Connected to inter-bodily multidimensional accounts of metaphor is the metaphor resistance phenomenon, only recently studied, and the various reasons why it happens.
For instance, people resist metaphors if they lack explanatory power or for a preference for alternative metaphorical concepts with respect to normative ones. However, without a comprehensive metaphor study it is not possible to know why some metaphors aren't picked up~\cite{gibbs2021we}. Thus, we aim to shed light on these theoretical studies to account for a multidimensional view of metaphor that can also reflect in a new strand of computational metaphor processing studies.

\subsection{Metaphor representation}

Computational representations of metaphor have leveraged structured resources such as MetaNet and Framester, which align conceptual metaphors with FrameNet frames and roles~\cite{gangemi2016framester}. The Amnestic Forgery Ontology further integrates MetaNet into Framester, providing a rich graph of source–target frame pairs, example sentences, and hierarchical relations among metaphors~\cite{gangemi_18}. Ontological formalisms based on the Blending Ontology\footnote{Available at \url{https://github.com/dersuchendee/BlendingOntology}.} 
encapsulate the four-space blending networks of CBT, permitting explicit encoding of input spaces, generic spaces, blends, and their mapping relations. Despite these advances, existing SWRL-based and rule-driven approaches often lack scalability and fail to account for tacit, context-dependent knowledge, limiting their applicability to open-ended or multimodal metaphor interpretation~\cite{Hamilton2021,Mitrovic2017}.

\subsection{Metaphor datasets and benchmarks}
Recent years have seen new dataset for computational metaphor processing. Among others, the VU Amsterdam Metaphor Corpus (VUA) has become a standard benchmark for metaphor detection~\cite{krennmayr2017vu}. However, such corpus only marks metaphoric tokens and does not specify the source and target domains behind each metaphor. To fill this gap, smaller domain-annotated datasets have emerged, such as the one by Gordon et al.~\cite{gordon2015corpus}, which annotates metaphorical tokens, source and target conceptual domains. A more comprehensive corpus is Metanet~\cite{dodge2013metanet}, a semantic wiki with conceptual metaphors that has been employed and extended in the Framester knowledge hub~\cite{gangemi2016framester}.
Lippolis et al. introduce the Balanced Conceptual Metaphor Testing Dataset (BCMTD), the first dataset that contains metaphors from the medical domain to test systems' generalizability~\cite{lippolis2025enhancing}.
At the same time, it is claimed~\cite{semino2004methodological} that in spite of the attention that metaphor has received over the centuries, and more recently within the cognitive paradigm, we still lack explicit and rigorous procedures for its identification and analysis, especially when one looks at authentic conversational data rather than decontextualized sentences.
Furthermore, more recently, doubts have been expressed about the legitimacy of extrapolating too readily from language to cognitive structure, and distinctions have been drawn between claims about whole linguistic communities or idealised native speakers, and claims about the minds of single individuals. 

\subsection{Theory-driven computational processing of metaphor}\label{theory-driven-processing}

Recent work in theory-driven metaphor processing integrates CMT and CBT into model architectures and annotation schemas. Mao et al.~\cite{mao2023metapro} and Tian et al.~\cite{tian2024theory} demonstrate that embedding theoretical constraints into training objectives improves metaphor detection performance. For interpretation, unsupervised and neural methods extract source and target domains or link attributes between them~\cite{shutova2017annotation,rosen2018computationally}, but typically depend on single-word annotations or pre-specified targets~\cite{wachowiak2023does}. Visual metaphor datasets such as MetaCLUE and ELCo provide multimodal challenges, yet public resources remain scarce~\cite{akula2023metaclue,yang2024elco}. In this context, neurosymbolic systems emerge. Logic-Augmented Generation (LAG) offers a promising paradigm by treating LLMs as reactive continuous knowledge graph generators, which convert text (and images) into structured semantic graphs and then enrich them with tacit knowledge to produce extended knowledge graphs that adhere to logical constraints~\cite{GANGEMI2025100859}. This hybrid approach has been explored in the work by Lippolis et al.~\cite{lippolis2025enhancing}, who showed that LLMs continue to struggle with metaphorical processing, especially domain specific, and multimodal inputs, despite seeing that a neurosymbolic system like LAG improves current metaphor performance. 

Furthermore, \citeauthor{lieto_delta_2025} \cite{lieto_delta_2025} recently presented a system called MET\textsuperscript{CL} able to perform metaphor generation and classification by applying a formal operationalization of the CBT. The core of MET\textsuperscript{CL} is a reasoning framework specialized in human-like commonsense concept combination: the Typicality-based Compositional Logic (T\textsuperscript{CL}) first presented in \cite{lieto_beyond_2019}, which is able to account for the composition of prototypical representations. The way MET\textsuperscript{CL} generates metaphor representations is grounded in the Categorization theory, but the system was also applied to the conceptual metaphors from MetaNet, which is based on Conceptual Metaphor Theory.


\subsection{Metaphors as speech acts}

Pragmatics, as the study of the meaning of linguistic signs in context—that is, in their actual use—deals primarily with implicit linguistic knowledge: the type of meaning it investigates, the pragmatic message, is not encoded in any direct way in the literal utterance. One can say something that literally means one thing while actually intending something entirely different. For instance, saying to someone on the subway, ``You’re standing on my foot'', means likely not wanting to describe the situation to them, but rather asking them to move. The utterance may contain only hints as to how the pragmatic meaning should be interpreted, such as a specific tone of voice, and it depends on a variety of extra-textual and extra-linguistic elements, including context, linguistic conventions, and socio-cultural norms. 

Utterances that contain metaphors can, of course, be described in pragmatic terms; however, pragmatic approaches to metaphor depend on the different approaches to the interpretation of metaphor itself. In that of two leading figures in the pragmatics literature and of SAT such as Grice  and Searle \cite{Grice1975-GRILAC-6, Searle1993-SEAM-3} there is the idea that the metaphorical interpretation presupposes and is derived from the literal interpretation through a ``decoding'' of a secondary level of meaning, as if metaphors and other figures of speech were some exceptional or supplementary use of language. The assumption of a clear-cut distinction between the literal interpretation and the metaphorical interpretation of an utterance is instead abandoned in the approach to metaphor of Sperber and Wilson \cite{Sperber1995-SPER, sperber1988representation}, the theorists of Relevance Theory (RT). RT, whose strength lies in its understanding of how language works through its emphasis on the inferential comprehension of non-literal meanings from contextual cues and assumptions, and on the cognitive principle that human communication aims at maximal relevance with minimal processing effort, treats metaphor as a pragmatic tool conveying implicatures and implicit evaluations.

The cognitive approach to the study of metaphors has been increasingly followed after the rise of CMT, and even in pragmatics, studies have focused their research on the many and varied effects that metaphor has on cognitive processes (e.g., \cite{Steen2008-STETPO-127, 10.3389/fpsyg.2023.1242888}). However, in computational pragmatics the field still appears to be very open: the pragmatic processing and understanding of metaphor—why it is used, in what context, and with what effect—remains only partially addressed, despite the advances of contextual neural models in recent years. Focusing exclusively on cognitive aspects may also obscure others, such as the performative dimension of utterances, which is instead well highlighted by SAT. Speaking of utterances as speech acts, in fact, means considering language and linguistic utterances not merely as expressions of mental operations, like articulations of thoughts, but as actions in themselves. As actions, while pursuing their speaker’s intentions, they produce effects in the world, whether intended or not, and carry out acts such as describing, ordering, pleading, or, through conventional formulas, marrying two people or sentencing someone.

A central aspect in explaining pragmatic meaning is the intention attributed to an utterance, the \textit{illocution}—that is, the way in which the speaker intends the literal sentence they utter, for example, saying something with the intention of making a request or giving an order. One of the main goals of computational pragmatics inspired by SAT—namely, studying how to automatically assign an \textit{illocutionary act} to the utterance intended to express it, framing this as a problem of context dependence—is complicated by several factors. First, there is the challenge of formalizing intentional, conventional, or otherwise contextual aspects that are extra-linguistic, along with all the choices that such formalization entails. Second, there is no deterministic relationship between clause types and illocutionary force: imperative clauses are not invariably commands, interrogative clauses are not always queries, and declarative clauses are not necessarily assertions. Finally, any illocutionary classification, however widely shared, will inevitably fall short of capturing the compositionality of the intentions at play in natural language. Depending on factors such as the power dynamics between speaker and addressee and their communicative goals, an utterance may emerge as a complex blend of illocutionary forces; and the same applies in the case of a ``metaphorical speech act''—understood here not as a distinct illocutionary category, but as a speech act that is in some way metaphorical.

Thus, as noted by Michelli, Tong and Shutova \cite{michelli2024framework}, the literature on metaphor intention is fragmented: there is still a lack of a systematic and comprehensive account of intentions behind metaphor use and an operationalised framework enabling the annotation of such intentions in linguistic data. However, explaining the communicative role of metaphors in terms of taxonomies of intentions, as they propose, may risk to perpetuate a classificatory game that views intentions as isolated. Another thing to note is that, although there is not a common notion of intention shared among all metaphor scholars, intentions are typically formalized as prior intentions, that is, as representations in the speaker’s mind of their communicative goals, especially in approaches that address metaphor from the perspective of models of Theory of Mind based on beliefs and intentions. As Gibbs \cite{gibbs1999intentions} points out, conceiving of intentions as individual mental states makes them opaque, since agents are not always aware of the causes of their behaviour; this, in turn, can lead to computational abstractions that lack real explanatory value for illocutionary intention.


Studying metaphors as speech acts, instead, means treating intention not as a mental state, but as a feature attributed to linguistic acts, in line with the philosophy of action outlined by Austin in his linguistic analyses \cite{austin1962things, Austin1961-AUSAPF-4}. 
This means that intentions, rather than being the reasons that speakers may provide when asked why they resorted to certain metaphors, are instead the intentions expressed by the utterance itself (thus interpreting an utterance as a command rather than a statement). Consequently, it is not immediately necessary to presuppose mental states beyond linguistic analysis. From a computational perspective, this entails modelling intention as an inferred property of the communication rather than as a presupposed mental condition—certainly a challenge, but also a potentially valuable contribution to the field of computational pragmatics.
Second, if metaphor-containing utterances are speech acts, they also produce effects on the communicative context, and part of which derive from the metaphor as a cognitive and stylistic device. The pragmatic effectiveness of a metaphorical speech act—how well the utterance serves the speaker’s communicative goals and fits the interlocutor’s sensitivity and context—is shaped by the stylistic tone conveyed through rhetorical figures and by the strength of the conceptual evocation created by the metaphor itself. Within a given illocutionary intention, a metaphor can either reinforce or attenuate that intention: a command may sound more polite, a request more engaging, or a threat more forceful; conversely, if the metaphor is perceived as inappropriate, its expressive or persuasive force may be diminished. This suggests that, beyond analysing illocutionary intention, it is also relevant to consider the perlocutionary level (to use Austin’s terminology, the \textit{perlocutionary}, the name for the speech act performed \textit{by} saying something) of a metaphorical speech act, which concerns the effect an utterance or image has on the listener (e.g., convincing, frightening, provoking thought). These effects can be cognitive—often more difficult to study—but also at a more immediate at the emotional-psychological level, since, as figures of speech, metaphors can aim at linguistic persuasion, including the evocation of emotional responses.


\section{How meaning is an onion: a proposal of three-dimensional meaning representation}
\label{sec:proposal}
\subsection{The layered perspective}
We will discuss now the main thesis of our research proposal and the three-dimensional meaning framework. It serves as a vocabulary that, once discussed within the community, can lay the groundwork for future computational implementations. The main aim for the framrwork is to provide, as an overall framework, the interpretation of metaphor as both a cognitive tool and a speech act.


\begin{figure}[htbp]
  \centering
  \scalebox{0.8}{%
  \begin{tabular}{ccc}
    \begin{subfigure}[b]{0.30\textwidth}
      \includegraphics[width=\linewidth,height=3cm, keepaspectratio]{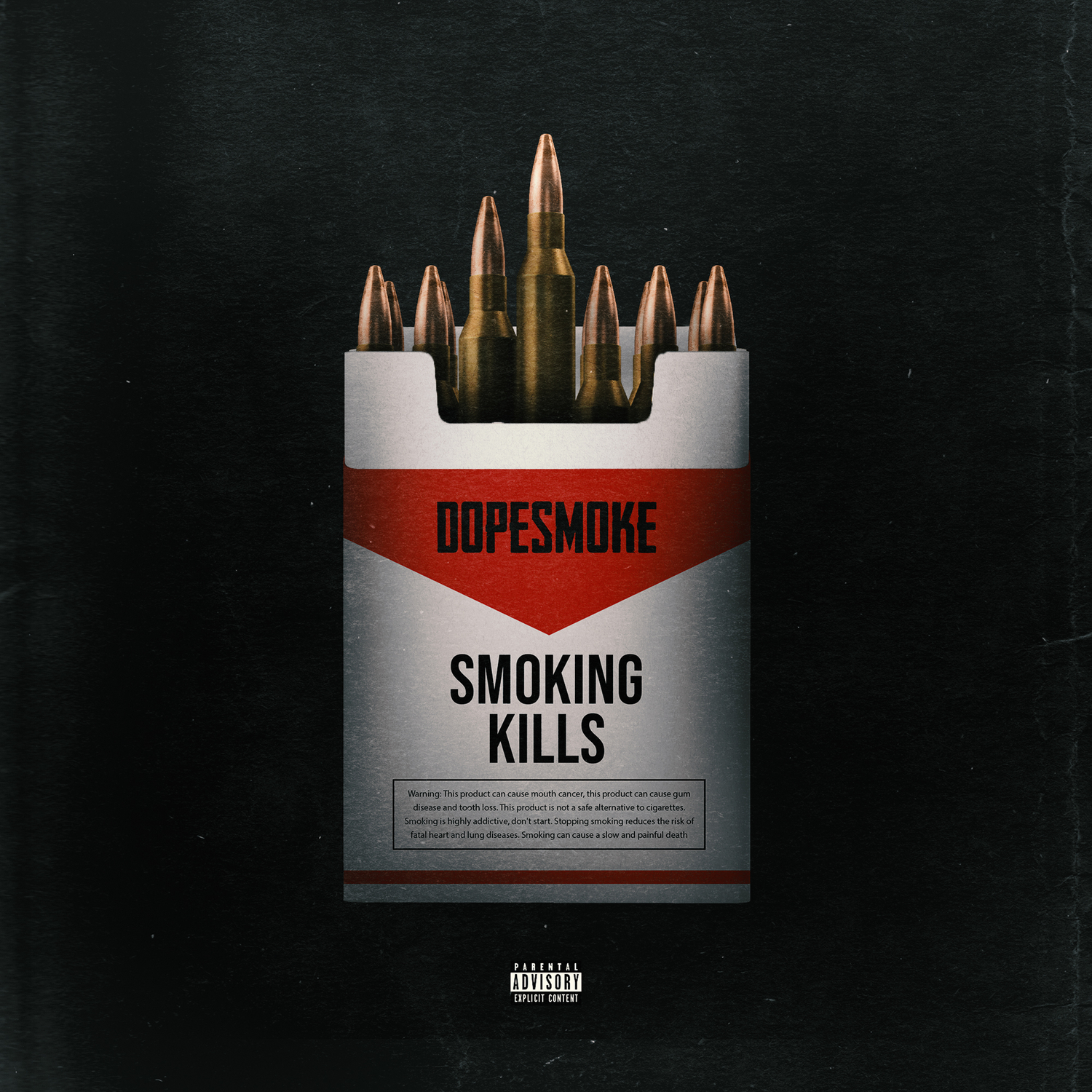}
      \caption{Image of the song ``Smoking kills'' by Dopesmoke.\footnote{\url{https://open.spotify.com/intl-it/track/6XL0aFfNf6PzhyzPddRArR}}}
    \end{subfigure}
    &
    \begin{subfigure}[b]{0.30\textwidth}
      \includegraphics[width=\linewidth]{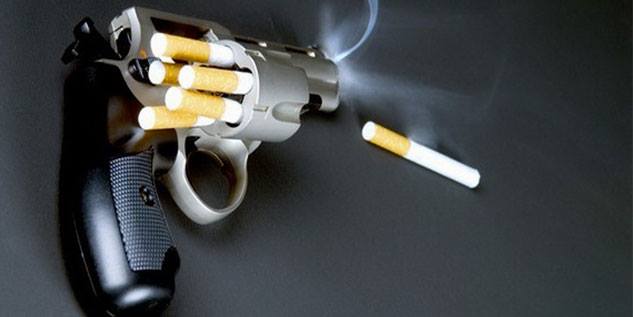}
      \caption{Anti-tobacco campaign from the online campaign \textit{No Smoke Revolution}.\footnote{\url{https://digitaladvocacycenterwher.com/en/gun-with-cigarette-bullets/}}}
      \label{nosmokerevolution}
    \end{subfigure}
    &
    \begin{subfigure}[b]{0.30\textwidth}
      \includegraphics[width=\linewidth]{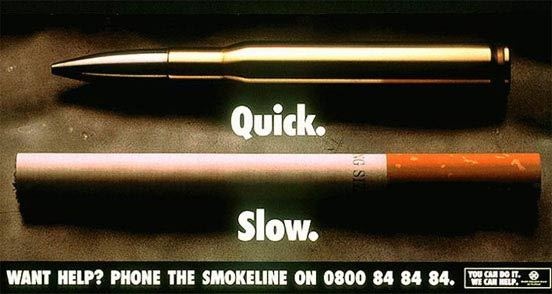}
      \caption{Ad by The Leith Agency for ASH UK (Dec 1994).\footnote{\url{https://adsspot.me/media/prints/ash-action-on-smoking-and-health-bullet-788f341dba7b}}}
    \end{subfigure}
  \end{tabular}%
  }
  \caption{Examples of anti-smoking imagery: the same metaphor can be conveyed in different contexts and with different goals.}
  \label{fig:smokingimg}
\end{figure}
\vspace{-1.5em}

\subsubsection{Layer 1: The external context layer} 

The conceptual content layer includes metadata of the communicative object we are analyzing: domain, provenance, connections to existing knowledge bases, etc. It also includes annotator metadata. 

\begin{figure}[h]
    \centering
    \includegraphics[width=0.6\linewidth]{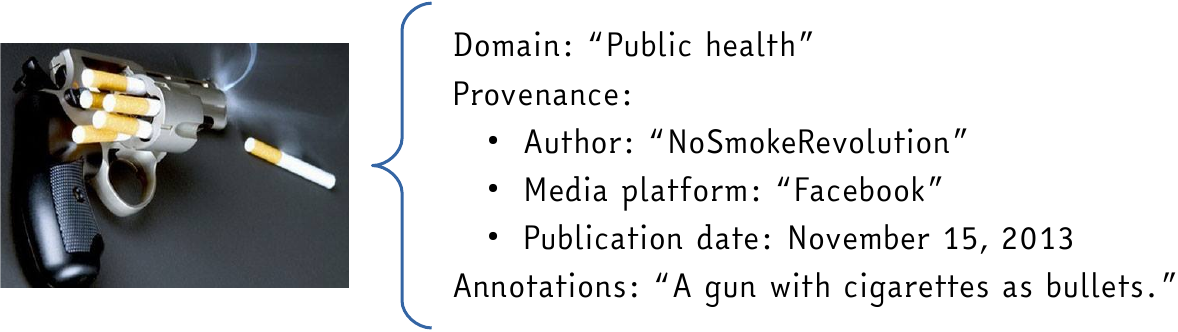}
    \caption{An example of ``Layer 1" analysis for Figure \ref{fig:smokingimg}b, showing the annotation of its content and related metadata.}
    \label{fig:layer-1-example}
\end{figure}

Consider the anti-smoking images in Figure \ref{fig:smokingimg}. First of all, this layer would include domain classification (public health advertising), provenance references (campaign organization, publication date, media outlet where it appeared), and frame connections linking to established conceptual metaphor databases such as Metanet's \textsc{harm is destruction}. The annotator metadata would capture the interpreter's geographical background, cultural context (attitudes toward smoking and firearms), and demographic information, recognizing that metaphor interpretation is inherently subjective and culturally situated. This foundational layer ensures that subsequent cognitive and pragmatic analyses can be properly contextualized within their social, temporal, and interpretive frameworks. Figure \ref{fig:layer-1-example} shows an example of annotations for the metaphorical image shown in Figure \ref{nosmokerevolution}.

\vspace{-1em}
\subsubsection{Layer 2: The conceptual combination layer}

This layer includes the cognitive context of the element, and in particular what enables the metaphorical mapping by taking into account only the two domains employed in the metaphors. Analyses that detect source and target concepts in a sentence or image are part of this layer, as well as the approaches that try to blend the two concepts to generate a representation of the metaphorical meaning. The representation of source and target can provide a rich description of the respective domains, including frame roles that describe related concepts and the semantics of their relations, including why they map (the blending principle).

The context of this layer is given at least by the annotation of source and target domains and is fully represented according to the Blending Ontology (see Section \ref{sec:related}) with the following elements: (i) Blendable (source and target domains); (ii) Blending (the blending principle); (iii) Blended (the actual blend).

To generate a representation of the metaphorical meaning that emerges from the blending, we need a computational mechanism to realize the combination of source and target.
Consider again one of the anti-smoking advertisement such as Figure \ref{fig:smokingimg}. The source domain is \textsc{shooting}, while the target domain is \textsc{smoking}. According to Blending Theory, the blendables are the two input spaces: the weapon frame (with roles like shooter, ammunition, target, harm) and the smoking frame (with roles like smoker, cigarettes, user, health consequences). The blending principle that enables this conceptual integration is \textit{lethality}—both bullets and cigarettes cause death, though through different temporal mechanisms. This creates an emergent blended space where cigarettes inherit the immediate, violent danger typically associated with ammunition, while smoking adopts the intentional, direct harm associated with shooting. The blended space produces the integrated concept where each cigarette becomes a bullet the smoker fires at themselves, creating a self-destructive cycle. Frame roles map systematically: the \textsc{shooter} maps onto the \textsc{smoker}, \textsc{ammunition} onto \textsc{cigarettes}, the \textsc{act of shooting} onto the \textsc{act of smoking}, and \textsc{fatal injury} onto \textsc{smoking-related disease}, unified by the overarching principle of \textit{lethality} that bridges the temporal gap between immediate and gradual self-harm.

Many attempts have been made to develop an algorithmic solution for conceptual combinations, each with its limitations. The Structure Mapping Engine (SME \cite{falkenhainer_structure-mapping_1989}) is a well-known computational approach that takes into account a broad description of source and target domains. Given the knowledge graphs representing the objects and the relations involved in the two domains, the SME finds isomorphic subgraphs that yeld the mappings. While powerful, the SME requires a rich, formal description of the two domains.
A less demanding approach, directly inspired by the Categorization theory, is MET\textsuperscript{CL} \cite{lieto_delta_2025}. Based on a cognitively inspired logic for conceptual combination, this system can automatically generate a prototypical representation of the metaphorical mapping (the blend) from the prototypical representations of source and target concepts. MET\textsuperscript{CL} consists in a three-step pipeline. The first step builds a structured representation of the metaphors to be analyzed, highlighting source and target. The second step generates a prototypical representation of both the source concept and the target concept. Each concept is represented by a prototype, i.e. a small set of typical features that can be automatically extracted from ConceptNet \cite{speer_conceptnet_2017}. The third step is the conceptual combination. The T\textsuperscript{CL} logic combines the source and target prototypes to generate an abstract representation of the metaphor. The combinations generated by MET\textsuperscript{CL} were generally accepted by human judges as capturing relevant aspects of the intended metaphorical meaning.

\begin{figure}[h]
    \centering
    \includegraphics[width=0.7\linewidth]{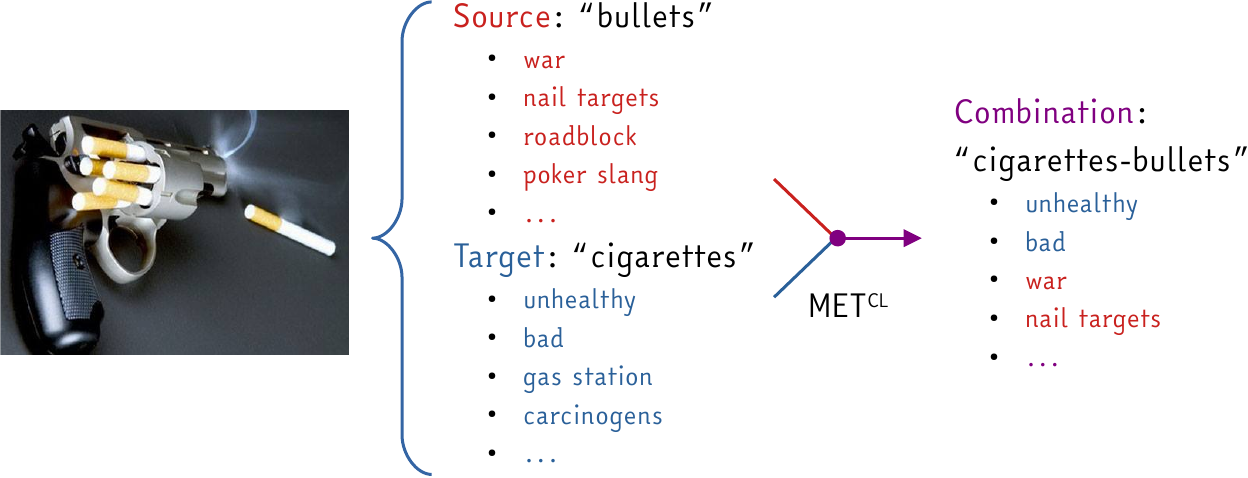}
    \caption{An example of ``Layer 2" analysis for Figure \ref{fig:smokingimg}b, showing the result of the conceptual combination performed by MET\textsuperscript{CL}.}
    \label{fig:layer-2-example}
\end{figure}

MET\textsuperscript{CL} can be seen as a tool for knowledge graph completion. Indeed, manually curated resources like MetaNet are inevitably incomplete and suffer from under-representation of the wide metaphor phenomenon\footnote{In particular, the MetaNet project is still under constant improvement and enrichment after more than a decade from its beginning. More information is available on \href{https://metanet.arts.ubc.ca/}{the website of the MetaNet Project}.}. The ability of the system to automatically generate a representation of a given metaphor allows to cover a wider spectrum of expressions. In Lieto et al.\cite{lieto_delta_2025}, LLMs were used to classify metaphorical sentences into the MetaNet ontology classes, showing the benefits of extending the ontology with the representations generated by MET\textsuperscript{CL}.
Our proposal for Layer 2 is to use a conceptual combination system like MET\textsuperscript{CL} as the reasoning mechanism. It can also be provided as a basis for the LAG-based approach by Lippolis et al. \cite{lippolis2025enhancing} (see Section \ref{sec:related}) that addresses multimodal metaphorical raw data and generates knowledge graphs based on implicit knowledge. Figure \ref{fig:layer-2-example} shows the MET\textsuperscript{CL} implementation of this layer.

\vspace{-1em}

\subsubsection{Layer 3: The pragmatic layer}

Compared with the first two layers, the third remains largely under‑investigated from a computational perspective. Our aim is to discuss a categorization that would allow for its effective implementation, also with a view to the composition and annotation of a dataset. The idea, in fact, is to propose a human annotation campaign on a dataset—such as one consisting of metaphors found in images, as in the example discussed so far—and then to assess the behavior of an automatic system on the same task.

 As seen in Section 2.5, works on the communicative role of metaphors in a pragmatic sense tend to explain it in terms of the intentions or discourse goals they are meant to achieve, proposing taxonomies of intentions \citep{michelli2024framework}, that often assumes intentions as mental states. For the analysis of intention as an illocutionary act, we reuse the standard categories of SAT, focusing on defining the role of metaphorical linguistic action\footnotemark{}. 

\footnotetext{The general classification of illocutionary acts (or language functions) has been a recurring topic in the philosophy of language literature, but here we are not concerned with defending a particular classification. We adopt an approach that would allow us to flexibly assess the composition of forces within an utterance containing a metaphor, focusing on the example of \textit{directives}—the linguistic category of requests or orders typically derived from imperative sentences and found across most languages and linguistic cultures \cite{Levinson1983-LEVP-16}. According to certain theoretical frameworks, such as those proposed by Popa-Wyatt, one could also distinguish primary and secondary illocutionary acts, sometimes nested within each other, as, for instance, in the case of an ironic act embedded inside a metaphorical one \cite{Popa-Wyatt2017-POPCFP}, but given the preliminary nature of this operational framework, we focus on annotating only the forces perceived as primary.}

As shown, a perlocutionary analysis aligns with evaluating not only what a metaphor means, but also what it does, treating utterances as actions that produce effects in the communicative context and in listeners, whether intended or not. We propose capturing these effects through a first, immediate psychological evaluation, and extend this approach across modalities, including the recognition of tone of voice in speech as a cue to pragmatic intention.

We therefore propose the following categorization:

\begin{itemize}
        \item \textbf{Attitude}: 
    The speaker’s evaluative stance toward the object of the metaphorical analogy (not toward individual discourse elements separately). This dimension reflects whether the attitude conveyed is positive, negative, or neutral, and can influence the tone and intention inferred from the utterance.

    \item \textbf{Illocutionary Act}: The act that expresses how speakers intend their literal utterance to be understood—thus representing the key element from which utterance's intention can be inferred.
    Following classifications such as Searle’s taxonomy, illocutionary acts can include assertive, directive, commissive, expressive, and declarative types. Communicative acts often involve a composite of multiple illocutionary forces, but our case study focuses specifically on elements that can be traced back to directive illocutionary acts as the principal pragmatic force perceived by annotators.
    
    \begin{itemize}
        \item \textbf{Directive Kind}: The definition of the specific type of directive illocutionary act according to a range of assertive \textit{force}. Directive acts are those in which the speaker attempts to get the addressee to perform an action, but this force can vary in intensity. Directives may take the form of requests, commands, suggestions, prohibitions, or pleas. We propose to model the type of directive along a continuum of assertiveness to capture this variation in illocutionary strength.
    \end{itemize}

    \item \textbf{Perlocutionary Act}: The act concerning the effects an utterance or image has on the listener/annotator, here analyzed as psychological or emotional. This includes both intended and unintended responses, which we propose to capture as annotators’ evaluations.
    \begin{itemize}
        \item \textbf{Perlocutionary Effect}: 
        The emotions evoked by the metaphor in the annotator, categorized according to the Emotion Frame Ontology \cite{de2024efo}. These emotional responses may vary and are optional: in some cases a metaphor may evoke no particular emotion.
        \item \textbf{Efficacy}: A measure (on a Likert scale from 1 to 5) of the metaphor's effectiveness, namely its capacity for persuasion relative to the presumed illocutionary intention, its appropriateness to the perceived context, and the annotator's personal sensitivity. 
    \end{itemize}
    
    \item \textbf{Other Contextual Elements}
    \begin{itemize}
        \item \textbf{Pragmatic/Visual Clues}: Indicators that suggest the speaker’s or communicator’s attitude toward the metaphor’s referent, e.g. use of specific colors.
        \item \textbf{Pragmatic/Visual Tone of Voice}: The overall communicative tone (e.g., humorous, dramatic, ironic), which contributes to how the metaphor and its communicative intent are perceived.
    \end{itemize}
\end{itemize}

\begin{figure}[h]
    \centering
    \includegraphics[width=0.7\linewidth]{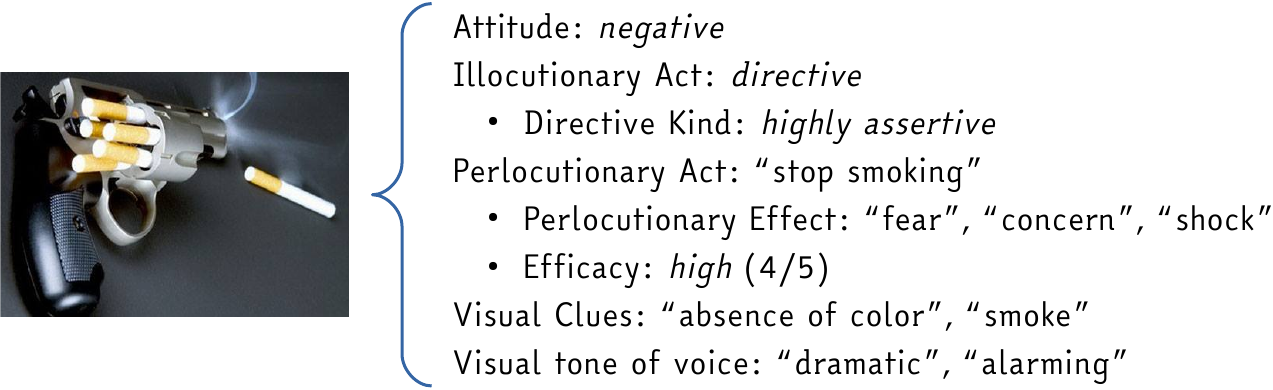}
    \caption{An example of ``Layer 3" analysis for Figure \ref{fig:smokingimg}b, showing its description according to pragmatics.}
    \label{fig:layer-3-example}
\end{figure}

As an example, consider again the image shown in Figure \ref{nosmokerevolution}. The speaker’s evaluative stance can easily be perceived by an annotator as negative toward the analogy's object: smoking is framed as a lethal threat and self-destructive act. This negative attitude is conveyed both conceptually (via the metaphor) and visually (via design choices). The utterance, although visual, can function as a directive illocutionary act because it is embedded in the context of anti-smoking public campaigns and visually represents a cigarette as a bullet. This metaphor implies a communicative function—discouraging smoking behavior by equating it with an act of self-inflicted violence. From this function, the speaker’s intention can be inferred as emerging from the pragmatic force and evaluative stance of the act itself, as to prevent smoking behavior by highlighting its deadly consequences through the conceptual mapping of \textsc{smoking is shooting oneself}. As for the directive kind, the discouraging attitude against smoking places the metaphorical act in a position that can be perceived by an annotator as the highly assertive end of the directive spectrum, referring to a prohibitive force similar to a command. This is achieved not through explicit verbal instruction but via the emotional and conceptual weight of the metaphorical image. The perlocutionary effects that can be experienced by an annotator can include emotions such as concern, discomfort, and shock as viewers process the visual equation of cigarettes with ammunition. The metaphor's efficacy can be perceived as potentially high due to the stark, unambiguous nature of the weapon imagery and its cultural associations with death and violence. Visual clues reinforce the negative attitude through the absence of color, and the conceptual mapping between shooting a gun and smoking a cigarette is underlined by the presence of smoke. The overall visual tone of voice is dramatic and alarming, designed to interrupt habitual thinking about smoking through visceral impact rather than rational argument. This example of analysis is outlined in Figure \ref{fig:layer-3-example}.


\vspace{-1em}

\section{Implications for computational metaphor processing and future work}
\label{sec:implications}
Our proposal holds several implications for computational metaphor processing. In this section, we outline them and propose future research that can implement the approach into an empirical framework, going back to the five problems described in Section \ref{sec:intro}.

\textbf{Datasets.}\label{subsub:dataset} Metaphor datasets should not only account for simple domain mappings within sentences, but also adopt the possibility to annotate conversational and multimodal data accounting for dimensions other than the cognitive one. To date, no single metaphor-related dataset annotates attitude, speech acts and clues of utterances, but initial effort is being made towards this step, for example in the representation of intentions \citep{michelli2024framework}. Furthermore, datasets concerning perlocutionary effects should also link to ontological resources about emotions. Additionally, no dataset yet accounts for the fact that meaning can be represented as a spectrum, thus future work should account for this conception of meaning.  Finally, employing diverse annotators is fundamental in the view employed in this paper, in order to obtain the first gold standard fine-grained pragmatic datasets in the field.

\textbf{Knowledge representation.}
Current ontological frameworks for metaphor representation, as described in Section \ref{sec:related}, have made important strides by encoding source–target mappings and blend spaces. However, they tend to treat metaphor as a phenomenon abstracted from discourse and pragmatic intent. In contrast, our proposal emphasizes a multi-layered model of meaning that integrates literal, conceptual, and pragmatic levels, requiring a representational shift that can, in future developments of this work, be ontologically formalized.
In fact, metaphor representation should move toward stratified ontological models that includes a pragmatic layer capturing the illocutionary and perlocutionary acts of metaphorical utterances, grounded in discourse context, communicative intent, and emotional effects.
Finally, such a layered representation can support updates or reinterpretations of metaphorical meaning in conversational settings; for instance, through reasoning mechanisms capable of tracking metaphor reinterpretation and resistance, grounded in community norms and individual differences.




\textbf{Possible application of the framework in computational metaphor processing.} The proposed framework has direct implications for computational processing. By aligning processing tasks with the three dimensions, we can design systems that generate, interpret, and apply metaphors in a more context-aware, human-aligned, and socially beneficial manner.
For instance, both the LAG approach \cite{lippolis2025enhancing} and the MET\textsuperscript{CL} \cite{lieto_delta_2025} approach can be employed in a complementary way to assess the conceptual content and combination layer for metaphor processing.
We argue that MET\textsuperscript{CL} and LAG are not only compatible but potentially integrable and mutually reinforcing. Specifically, LAG could leverage the prototypical property lists provided by MET\textsuperscript{CL} to weight RDF triples, yielding more precise definitions and reducing LLM hallucinations. Conversely, MET\textsuperscript{CL} could exploit LAG's capacity to identify missing frames in MetaNet or to align typical features with ontological categories, thus enabling a shared RDF layer for interoperability. For instance, LAG might detect objects in an image, while MET\textsuperscript{CL} could suggest which visual attributes instantiate their prototypical properties.
For what concerns the pragmatic layer, the lack of datasets currently hinders computational implementation (see Section\ref{subsub:dataset}), starting from a gold-standard annotated dataset, but LLM-based approaches such as LAG can be used to extract pragmatic features, which can in turn be analyzed.

For what concerns \textit{metaphor generation}, current methods often rely on shallow associations or fixed templates, leading to output that lacks conceptual coherence or pragmatic fit. By incorporating the layered representation we propose, generation systems—particularly those embedded in interactive environments such as chatbots or digital companions—can produce metaphors that are not only structurally sound, but also functionally appropriate to the communicative context. Likewise, \textit{metaphor-aware recommender systems} could translate figurative queries (e.g., ``a sofa with a cozy vibe'') into concrete product attributes, letting users find items that match their affective intent while keeping the system’s reasoning transparent.


\textit{Metaphor understanding}, too, can benefit from the layered representation we propose. By encoding both conceptual mappings and pragmatic implications, systems can disambiguate metaphorical usage more accurately, especially in real-life settings. LLMs, when coupled with structured semantic representations, may serve as effective metaphor interpreters. In such a neurosymbolic setup, the literal surface form is parsed, mapped to conceptual domains, and then interpreted through pragmatic filters reflecting the speaker's goal, emotional tone, and discourse situation. This facilitates better performance in metaphor detection, paraphrasing, and explanation tasks, especially in underexplored genres like dialogue, narratives, or scientific texts.

\textbf{Metaphors for social good.} Metaphors are not neutral; they carry social, cultural and emotional weight, and using or detecting them responsibly has implications across several socially relevant domains (see Section \ref{sec:related}), for instance employing metaphor-aware computational tools in education, science communication, doctor-patient interactions and hate speech. 

\section{Limitations and conclusion}
\label{sec:conclusion}
Metaphors are cognitive tools deeply embedded in human reasoning and communication. In this work, we assess current problems in metaphor processing and propose a three-layered framework for metaphor processing, encompassing conceptual and pragmatic implicit knowledge together. This onion-like stratification allows computational models to capture not only what metaphors \textit{mean} but what they \textit{do}, illuminating both their cognitive structure and their communicative function.
Although the operational framework is currently theoretical, we show how it can serve both as a fruitful direction for future research and as a foundation for subsequent empirical work on this issue, for instance, for the design of computational systems that are metaphor-aware, context-sensitive, and socially responsible. By formalizing metaphor as a structured, multi-dimensional phenomenon, we lay the foundation for new datasets, evaluation metrics, and neurosymbolic methods that better reflect the richness of metaphor in natural discourse. The main current limitation of this framework is the lack of computational implementation. It is also necessary to develop a robust and shared line of research that integrates CBT and pragmatics theories on metaphor interpretation, accompanied by the discussion of a common vocabulary for the construction of future datasets, and, on the computational side, by a coherent formalization of pragmatic aspects such as intentional and contextual factors. Looking ahead, we envision extending this framework to new modalities and domains and implementing it in computational experiments.

\begin{acknowledgments}
\footnotesize

This work was supported by the PhD scholarship ``Discovery, Formalisation and Re-use of Knowledge Patterns and Graphs for the Science of Science'', funded by CNR-ISTC through the WHOW project (EU CEF programme - grant agreement no. INEA/CEF/ICT/ A2019/2063229).

\end{acknowledgments}

\section*{Declaration on generative AI}
\footnotesize

 In the preparation of this work, the authors used GPT-4 and Grammarly in order to: Grammar and spelling check. After using these tools, the authors reviewed and edited the content as needed and take full responsibility for the publication’s content. 
\bibliography{sample-ceur}

\appendix



\end{document}